\documentclass[pdflatex,sn-mathphys]{sn-jnl}
\normalbaroutside

\usepackage{amsmath,amssymb,amsfonts}
\usepackage{lineno}
\usepackage{graphicx}
\usepackage{textcomp}
\usepackage{multirow}
\usepackage{colortbl}
\usepackage{hhline}
\usepackage{float}
\usepackage{adjustbox}
\usepackage[T1]{fontenc}
\usepackage{textcomp}
\usepackage{lmodern}
\usepackage{ulem}
\def\BibTeX{{\rm B\kern-.05em{\sc i\kern-.025em b}\kern-.08em
    T\kern-.1667em\lower.7ex\hbox{E}\kern-.125emX}}
    
\usepackage[xindy]{glossaries}

\newacronym{acaces}{ACACES}{Advanced Computer Architecture and Compilation for high-performance Embedded Systems}
\newacronym{acm}{ACM}{Association for Computing Machinery}
\newacronym{aig}{AIG}{AND-inverter graph}
\newacronym{ai}{AI}{artificial intelligence}
\newacronym{ann}{ANN}{artifical neural network}
\newacronym{asic}{ASIC}{application-specific integrated circuit}
\newacronym{asr}{ASR}{automatic speech recognition}
\newacronym{avx}{AVX}{advanced vector extensions}
\newacronym{aws}{AWS}{Amazon Web Services}
\newacronym{bat}{BAT}{Baidu, Alibaba, Tencent}
\newacronym{bert}{BERT}{bidirectional encoder representations from transformers}
\newacronym{bfloat16}{bfloat16}{brain floating point 16-bit}
\newacronym{bic}{BIC}{bit clear}
\newacronym{blas}{BLAS}{basic linear algebra subprograms}
\newacronym{bnn}{BNN}{binary neural network}
\newacronym{bram}{BRAM}{block RAM}
\newacronym{brnn}{BRNN}{bidirectional recurrent neural network}
\newacronym{cal}{CAL}{Computer Architecture Letters}
\newacronym{cifar}{CIFAR}{Canadian Institute For Advanced Research}
\newacronym{clb}{CLB}{configurable logic block}
\newacronym{clip}{CLIP}{Contrastive Language-Image Pre-training}
\newacronym{cnn}{CNN}{convolutional neural network}
\newacronym{cpu}{CPU}{central processing unit}
\newacronym{cudnn}{cnDNN}{NVIDIA CUDA Deep Neural Network}
\newacronym{dag}{DAG}{directed acyclic graph}
\newacronym{darpa}{DARPA}{Defense Advanced Research Projects Agency}
\newacronym{dbn}{DBN}{deep belief network}
\newacronym{des}{DES}{data encryption standard}
\newacronym{dl}{DL}{deep learning}
\newacronym{dml}{DML}{deep metric learning}
\newacronym{dnn}{DNN}{deep neural network}
\newacronym{dram}{DRAM}{dynamic RAM}
\newacronym{dsp}{DSP}{digital signal processor}
\newacronym{dvs}{DVS}{dynamic vision sensor}
\newacronym{eda}{EDA}{electronic design automation}
\newacronym{edram}{eDRAM}{embedded \gls{dram}}
\newacronym{eie}{EIE}{efficient inference engine}
\newacronym{elu}{ELU}{exponential linear unit}
\newacronym{fc}{FC}{fully connected}
\newacronym{fsl}{FSL}{few-shot learning}
\newacronym{fft}{FFT}{fast fourier transform}
\newacronym{fifo}{FIFO}{first in first out}
\newacronym{flopoco}{FloPoCo}{FLOating-POint COres}
\newacronym{flops}{FLOPS}{floating-point operations per seconds}
\newacronym{flop}{FLOP}{floating-point operations per seconds}
\newacronym{fma}{FMA}{fused multiply add}
\newacronym{fpga}{FPGA}{field programmable gate array}
\newacronym{fps}{FPS}{frames per second}
\newacronym{fpu}{FPU}{floating-point unit}
\newacronym{fp}{FP}{floating-point}
\newacronym{fsm}{FSM}{finite-state machine}
\newacronym{gan}{GAN}{generative adversarial network}
\newacronym{gdpr}{GDPR}{General Data Protection Regulation}
\newacronym{gemm}{GEMM}{general matrix multiply}
\newacronym{gflops}{GFLOPS}{giga \gls{flops}}
\newacronym{gmafia}{GMAFIA}{Google, Microsoft, Apple, Facebook, Intel, Amazon}
\newacronym{gops}{GOPS}{giga operations per second}
\newacronym{gpgpu}{GPGPU}{general purpose \gls{gpu}}
\newacronym{gpu}{GPU}{graphics processor unit}
\newacronym{hdl}{HDL}{hardware description language}
\newacronym{hipeac}{HiPEAC}{High Performance and Embedded Architecture and Compilation}
\newacronym{hls}{HLS}{high-level synthesis}
\newacronym{hobflops}{HOBFLOPS}{hardware optimized bit-sliced floating-point operators}
\newacronym{hpm}{HPM}{high performance mobile}
\newacronym{ibm}{IBM}{International Business Machines}
\newacronym{ic}{IC}{integrated circuit}
\newacronym{ieee}{IEEE}{Institute of Electrical and Electronic Engineers}
\newacronym{iet}{IET}{Institution of Engineering and Technology}
\newacronym{ifm}{IFM}{input feature map}
\newacronym{ilp}{ILP}{instruction-level parallelism}
\newacronym{ilsvrc}{ILSVRC}{ImageNet large scale visual recognition challenge}
\newacronym{iob}{IOB}{input output buffer}
\newacronym{iot}{IoT}{internet of things}
\newacronym{ip}{IP}{intellectual property}
\newacronym{isa}{ISA}{instruction set architecture}
\newacronym{iso}{ISO}{International Organization for Standardization}
\newacronym{isscc}{ISSCC}{International Solid-State Circuits Conference}
\newacronym{jpeg}{JPEG}{Joint Photographic Experts Group}
\newacronym{knn}{KNN}{K-nearest neighbor}
\newacronym{lfw}{LFW}{Labeled Faces in the Wild}
\newacronym{lpddr}{LPDDR}{low-power double data rate}
\newacronym{lstm}{LSTM}{long short term memory}
\newacronym{lut}{LUT}{look up table}
\newacronym{lzc}{LZC}{leading zeros count}
\newacronym{lzoc}{LZOC}{Leading Zero or One Count}
\newacronym{mac}{MAC}{multiply-accumulate}
\newacronym{mcmk}{MCMK}{multi-channel, multi-kernel convolution}
\newacronym{mec}{MEC}{memory-efficient convolution}
\newacronym{mit}{MIT}{Massachusetts Institute of Technology}
\newacronym{mlp}{MLP}{multilayer perceptron}
\newacronym{ml}{ML}{machine learning}
\newacronym{mnist}{MNIST}{Modified National Institute of Standards and Technology}
\newacronym{mog}{MoG}{Mixture of Gaussians}
\newacronym{mosfet}{MOSFET}{metal oxide semi-conductor field effect transistor}
\newacronym{mux}{MUX}{multiplexer}
\newacronym{nan}{NAN}{not-a-number}
\newacronym{ncs}{NCS}{neural compute stick}
\newacronym{nlp}{NLP}{natrual language processing}
\newacronym{nn}{NN}{neural network}
\newacronym{npu}{NPU}{network processing unit}
\newacronym{nre}{NRE}{non-recurring engineering}
\newacronym{ofm}{OFM}{output feature map}
\newacronym{ops}{OPS}{operations per second}
\newacronym{osl}{OSL}{one-shot learning}
\newacronym{pasm}{PASM}{parallel accumulate shared MAC}
\newacronym{pas}{PAS}{parallel accumulate and store}
\newacronym{pcb}{PCB}{printed circuit board}
\newacronym{pcie}{PCIe}{peripheral component interconnect express}
\newacronym{pe}{PE}{processing element}
\newacronym{pflops}{PFLOPS}{peta \gls{flops}}
\newacronym{prelu}{PRELU}{parameteric rectified linear unit}
\newacronym{qcd}{QCD}{quantum chromodynamics}
\newacronym{ram}{RAM}{random access memory}
\newacronym{rbm}{RBM}{restricted Boltzmann machine}
\newacronym{relu}{ReLU}{rectified linear unit}
\newacronym{rl}{RL}{reinforcement learning}
\newacronym{rnn}{RNN}{recurrent neural network}
\newacronym{roi}{ROI}{return on investment}
\newacronym{ros}{ROS}{robot operating system}
\newacronym{rtl}{RTL}{register transfer logic}
\newacronym{sat}{SAT}{satisfiability}
\newacronym{sdc}{SDC}{Synopsys design constraint}
\newacronym{sfi}{SFI}{Science Foundation Ireland}
\newacronym{sgd}{SGD}{stochastic gradient descent}
\newacronym{shave}{SHAVE}{streaming hybrid architecture vector engine}
\newacronym{simclr}{SimCLR}{a Simple framework for Contrastive Learning of visual Representations}
\newacronym{simd}{SIMD}{single instruction multiple data}
\newacronym{sipp}{SIPP}{streaming image processing pipeline}
\newacronym{slide}{SLIDE}{sub-linear deep learning engine}
\newacronym{snarc}{SNARC}{Stochastic Neural Analog Reinforcement Calculator}
\newacronym{soi}{SOI}{silicon on insulator}
\newacronym{sop}{SOP}{sum-of-products}
\newacronym{sram}{SRAM}{static RAM}
\newacronym{svhn}{SVHN}{street view house numbers}
\newacronym{svm}{SVM}{support vector machine}
\newacronym{swar}{SWAR}{\gls{simd} within a register}
\newacronym{taco}{TACO}{Transactions on Architecture and Code Optimization}
\newacronym{tcl}{TCL}{tool command language}
\newacronym{tco}{TCO}{total cost of ownership}
\newacronym{tflops}{TFLOPS}{tera \gls{flops}}
\newacronym{tnn}{TNN}{ternary neural network}
\newacronym{tops}{TOPS}{tera operations per second}
\newacronym{tpu}{TPU}{tensor processing unit}
\newacronym{tvlsi}{TVLSI}{Transactions on Very Large Scale Integration}
\newacronym{twn}{TWN}{ternary weight network}
\newacronym{vfp}{VFP}{vector floating point}
\newacronym{vhdl}{VHDL}{very high speed integrated circuits hardware description language}
\newacronym{vliw}{VLIW}{very long instruction word}
\newacronym{xdc}{XDC}{Xilinx design constraint}
\newacronym{xpe}{XPE}{Xilinx power estimator}
\newacronym{zsl}{ZSL}{zero-shot learning}

\jyear{2021}%

\theoremstyle{thmstyleone}%
\theoremstyle{thmstyletwo}%

\theoremstyle{thmstylethree}%

\begin{document}

\title[Feature Representation in Pretrained Deep Metric Embeddings]{Feature Representation in Pretrained Deep Metric Embeddings}


\author*[1]{\fnm{Ryan} \sur{Furlong}}\email{ryan.furlong@itcarlow.ie}

\author[1]{\fnm{Vincent} \sur{O'Brien}}\email{vincent.obrien@setu.com}

\author[1]{\fnm{James} \sur{Garland}}\email{james.garland@setu.com}

\author[2]{\fnm{Daniel} \sur{Palacios-Alonso}}\email{daniel.palacios@urjc.es}

\author[2]{\fnm{Francisco} \sur{Dominguez-Mateos}}\email{francisco.dominguez@urjc.es}

\affil*[1]{\orgdiv{Department of Electronic Engineering and Communications}, \orgname{South East Technological University SETU}, \orgaddress{\street{Kilkenny Road}, \city{Carlow}, \country{Ireland}}}

\affil[2]{\orgdiv{Bioinspired Systems \& Applications Group Escuela T\'ecnica Superior de Ingenier\'ia Inform\'atica}, \orgname{Universidad Rey Juan Carlos}, \orgaddress{\street{28933 Móstoles}, \city{Madrid}, \country{Spain}}}


\abstract{In \gls{dml}, high-level input data are represented in a lower-level representation (embedding) space, such that samples from the same class are mapped close together, while samples from disparate classes are mapped further apart. In this lower-level representation, only a single inference sample from each known class is required to discriminate between classes accurately. The features a DML model uses to discriminate between classes and the importance of each feature in the training process are unknown. To investigate this, this study takes embeddings trained to discriminate faces (identities) and uses unsupervised clustering to identify the features involved in facial identity discrimination by examining their representation within the embedded space. This study is split into two cases; intra class sub-discrimination, where attributes that differ between a single identity are considered; such as beards and emotions; and extra class sub-discrimination, where attributes which differ between different identities/people, are considered; such as gender, skin tone and age. In the intra class scenario, the inference process distinguishes common attributes between single identities, achieving 90.0\% and 76.0\% accuracy for beards and glasses, respectively. The system can also perform extra class sub-discrimination with a high accuracy rate, notably 99.3\%, 99.3\% and 94.1\% for gender, skin tone, and age, respectively.}




\maketitle

\section{Introduction}\label{sec1}

Deep metric learning has proven to be highly adept at processing non-linear data, which in turn has led to the successful development of \gls{fsl} \cite{wang2020generalizing}, \gls{osl} \cite{santoro2016one} and \gls{zsl} \cite{10.1145/3293318} techniques. When applied to images, deep metric learning models first learn discriminative features from the images and then create embeddings representing the images. These embeddings are created such that when measured using a relatively simple Euclidean distance,  the embeddings of similar images are closer, and the embeddings of dissimilar images are farther apart. It remains to be seen which features beyond trained instances are learned by \gls{dml} models and whether such features can be utilised to perform sub-discrimination.

Regarding this, given labels, a deep metric learning model learns a new representation of input data. The model will learn distinguishing features between classes over many training iterations; however, the features learned between labelled classes are often unknown. For example, embeddings trained to distinguish between different identities would be provided with training labels whereby each unique identity holds a unique label. These labels will only inform the model that person A is not person B. During training, the model will learn features that best differentiate person A from person B and correlate person A to person A. The features learned by the model to perform these distinctions are, for the most part, unknown to us. This work aims to identify what features a specific deep metric facial recognition model learns during training and which features are most valuable to the discrimination ability of the model.
Consequently, if these valuable features can be accurately identified, can they be used to perform sub-discrimination? Sub-discrimination is the ability to perform discrimination on features within and between classes, features that are not labelled during the training process. We consider sub-discrimination a form of \gls{zsl} as the network is never trained to perform this type of discrimination. 

To answer these questions, we perform an experimental study that takes a pretrained network designed to discriminate faces and apply our sub-discrimination technique to differentiate between facial attributes, such as gender, age, and skin tone, without additional training. The attributes are split into extra and intra class. Extra class attributes are defined as distinguishable facial features between different identities, e.g. age, gender, and skin tone. Intra class attributes are distinguishable within a single identity, namely the presence of beards or glasses and the expression of different moods and emotions, e.g. happy, angry, sad, and neutral. This experimental study aims to provide insight into the representation and structure of features within \gls{dml} embeddings.

The rest of the paper is organised as follows: Section~\ref{sec2} provides background material and discussions on related work, such as N-shot learning and deep metric learning. Section~\ref{sec3} describes the experimental set-up for performing discriminative studies of embeddings from high-accuracy models. Section~\ref{sec4} details the results and an evaluation of the study. Section~\ref{sec5} describes the application of the results to perform reliable sub-discrimination on unseen images. Finally, Section~\ref{sec6} discusses future work and Section~\ref{sec7} provides a conclusion.

\section{Background \& Related Works}\label{sec2}
The field of machine learning has seen enormous growth in numerous application spaces due to the ability of machine learning models to outperform traditional approaches. The success of deep learning in areas such as computer vision, natural language processing and speech recognition has been principally due to the ability to leverage large amounts of data to develop models capable of meaningful representation \cite{gorriz2020artificial}. However, accessing large high-quality datasets is not always possible, and the lack of such datasets can present a significant barrier to developing machine learning models. This issue is not an isolated case of deep learning; other research areas suffer from the same restriction \cite{tan2015benchmarking,hu2016dvs,hasrul2012human}. With this in mind, methods aimed at solving the problems associated with limited data availability have emerged; namely, few-shot learning \cite{wang2020generalizing}, one-shot learning \cite{santoro2016one}, and zero-shot learning \cite{10.1145/3293318}. These learning methods are commonly utilised in computer vision tasks, where employing an object categorisation model still gives appropriate results even with zero, one or few training samples. These methods may use generative \cite{10.5555/3045390.3045551,Lake2015} or discriminative models \cite{VanesaSancho2011,Fan2014, Lin2015,Parkhi2015} to achieve their goal. This work studies one of the most common discriminative approaches, deep metric learning, where a model generates a vector in a low dimensional space with the help of a similarity metric.

\gls{dml} proposes to train an artificial neural network based on a non-linear feature encoder that embeds the extracted features that are semantically similar, close to one another, and maps distinct features further away from each other using an appropriate distance metric. \gls{dml} improves on the linear constraints of metric learning approaches, whereby kernel tricks \cite{Cortes1995} are required for the non-linear classification task. The \gls{dml} approach uses the non-linear activation functions of neural networks to solve the problem of nonlinearity. \gls{dml} approaches have achieved exceptional results on various tasks ranging from face verification, and recognition to three-dimensional (3D) modelling, a broad review of these studies can be seen in \cite{Kulis2012}. Specifically, with regards to image recognition, it can be seen that on several hard, fine-grained datasets \cite{Khosla2011, Nilsback2008, Parkhi2012, Russakovsky2012} where the examples are difficult to distinguish, e.g. images of objects from the same class, deep metric learning classification can eclipse the state-of-the-art \cite{Karlinsky2019}.

The \gls{dml} objective can be viewed as a solution to the one-shot learning problem, whereby one instance of a class is required to classify a new unknown instance from the same class. In this study, we manipulate the one-shot learning capabilities of embeddings generated by \gls{dml} models to perform near zero-shot learning, where salient attribute information unknown during the training process is used to sub-discriminate known classes.

Early work by \cite{Hadsell2006} is a strong motivation behind this experimental study. The authors demonstrate a mapping test of horizontally translated \gls{mnist} \cite{MNIST1998:LeCun} digits to a 2D output manifold. This manifold exhibits clustering with respect to the translations despite the system not being trained to do so. The network is trained to map digits of the same class close to one another and digits of different classes further apart; however, the authors note that coherent clusters are formed with respect to the five translations, and each cluster is well organised for class discrimination. This clustering can be viewed as a result of the coherence produced between features when the system learns a representation of the input data. More recently, \gls{zsl} approaches focus on creating novel classifiers that are generalised enough to accurately classify unseen classes from many independent datasets. A recent approach known as \gls{simclr} \cite{Chen2020} displays the same coherence demonstrated by \cite{Hadsell2006}. In their work, the authors conduct the pretraining in an unsupervised manner, whereby the network is given no labelled instances of the training data. The network is trained to map an image close to an augmented version of the same image and further away from images that are not augmented versions of themselves. The \gls{simclr} model achieves state-of-the-art classification results on the ImageNet dataset and even outperforms a strong supervised baseline despite the lack of class labels. The \gls{simclr} model can discriminate between disparate classes due to the coherence displayed in the new latent space learned by the network during the training stage.

Another recent example, \gls{clip} \cite{Radford2021}, has shown remarkable zero-shot classification results on numerous datasets. During training, \gls{clip} takes images scraped from the Internet and passes them through an image encoder. The text associated which each image, generally in the form of a caption, tag etc., is passed through a text encoder. The model is trained to map similar image embeddings and word embedding close to one another and dissimilar embeddings further apart. At inference, new unseen images are passed through the pretrained model and are represented as embeddings; labels are taken from the specific classification task and are also represented as embeddings. Zero-shot classification is achieved by using the label embeddings as prompts and classifying the images based on the prompt it lies closest to. \gls{clip} represents the current state-of-the-art for zero-shot classification and even outperforms a fully supervised linear classifier.
In contrast to the approach taken by \cite{Radford2021}, whereby the authors seek to create a universal zero-shot model, the work we propose focuses on examining the \gls{zsl} capabilities of pretrained \gls{dml} models. Our method utilises clustering to manipulate the coherence between features within the structure of embeddings generated by \gls{dml} models to identify the important features learned during training which can then be used to sub-discriminate between classes. We restrict our study to a DML face recognition model. However, if all \gls{dml} models exhibit similar coherence, then our method can be tailored to specific applications by substituting the pretrained models.

In \cite{Jain2018}, the authors propose a facial clustering method, whereby the authors use various models such as ResNet 50 initialised with weights pre-trained on the ImageNet \cite{deng2009imagenet} dataset for feature extraction, before scaling the data, performing dimensionality reduction, and finally clustering the data for the facial recognition task. Similarly, we take embeddings generated by a pre-trained facial recognition model but apply clustering to identify what facial attributes are represented within the embedding structure before using this inherent information to perform near zero-shot facial attribute classification. Consequently, our proposed work focuses on sub-discrimination (classification of facial attributes), not discrimination (classifying identities).
\section{Experimental Study}\label{sec3}

This study uses clustering to analyse embeddings generated for facial recognition to observe the feature information contained in the embedding structure. Additionally, the study investigates whether a hierarchy of representation exists among these features, indicating if certain features have a more robust representation in the embedding structure.

{\subsection{Deep Metric Learning Model}\label{subsec:dmlm}}
The model chosen for the experiment is the \textit{dlib face recognition ResNet model v1} contained in the Dlib library \cite{King2009}. This model was selected as it supplies an open source, easy-to-use facial recognition \gls{cnn} with a classification accuracy of 99.38\% on the standard \gls{lfw} face recognition dataset \cite{Liu2015}. The \textit{dlib face recognition ResNet model v1} is based on the ResNet 29 architecture \cite{7780459}. The model's creator trained it using approximately three million faces, and 7485 individual identities; any overlap with the \gls{lfw} dataset was avoided. The work of \cite{King2009} specifies that the datasets used to train the network include the Face Scrub dataset \cite{Hong-Wei}, the VGG dataset \cite{Parkhi2015} and images sourced from the Internet. The model performs facial recognition by mapping images of faces to a 128-dimensional space where images of the same identity are mapped near each other and images of different identities are mapped far apart in the new embedding space. \cite{King2009} states that the network training started with randomly initialised weights and used a metric loss function that tries to project all identities into non-overlapping clusters of distance threshold radius 0.6. When using a distance threshold of 0.6, the model received its highest accuracy of 99.38\% on the standard \gls{lfw} face recognition benchmark.

\begin{figure}[t]
\centering
\includegraphics[scale=0.6]{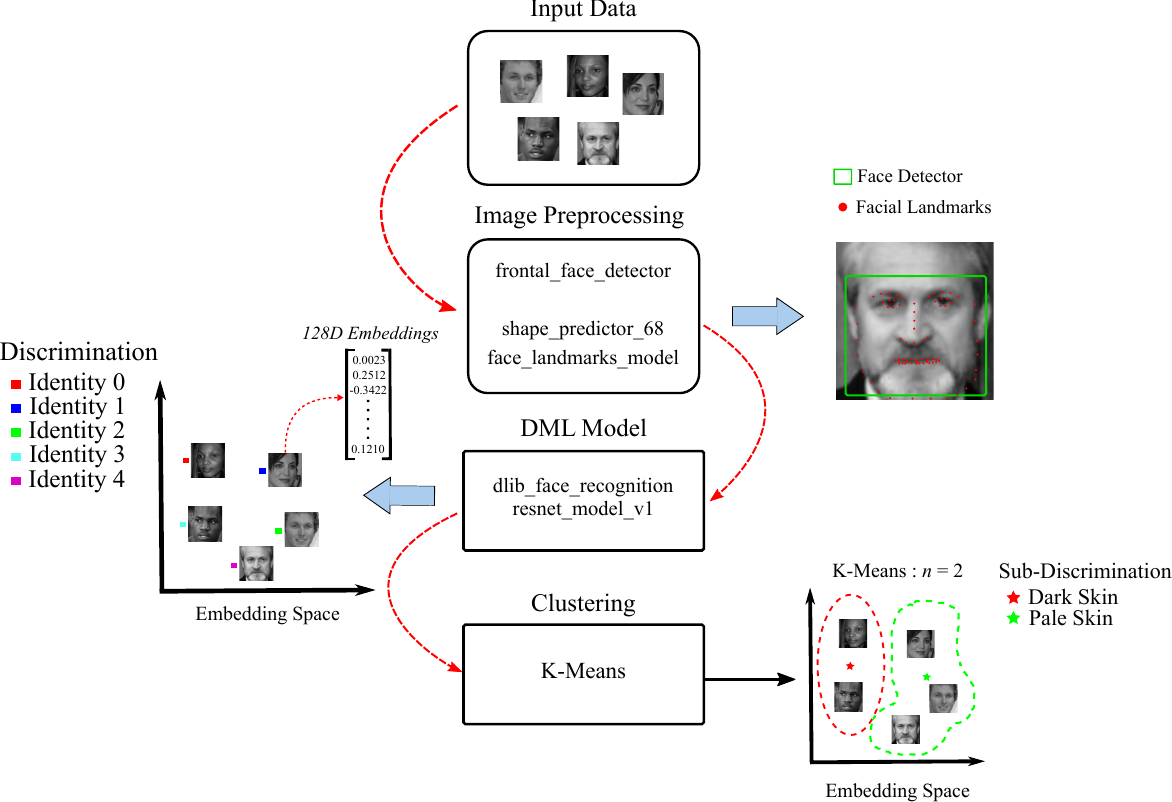}
\caption{Flowchart of the experimental process.}
\label{fig:figure1}
\end{figure}

The facial recognition task can be achieved by applying a discriminative classification algorithm such as \gls{knn} onto the embeddings without being conditioned about the number of classes or, in our case, the number of identities. Consequently, it is possible to perform facial recognition using only one example, achieving one-shot learning.
In this study, we adopt the K-Means and \gls{mog} algorithms to gain a better insight into attribute representation within the embedding structure. This work shows that the model learns specific facial attributes during training as K-Means successfully form coherent clusters where salient intra and extra class attributes can be accurately classified. As a result of this, embeddings generated for one purpose, in our case, facial recognition, can find an additional application in sub-discriminating intra and extra class attributes through no additional retraining of the \gls{dml} model.  
   
{\subsection{Experimental Procedure}\label{subsec:experiment procedure}}
Several tests were undertaken to examine what features are learned by this specific \gls{dml} model during training. For these tests, datasets with specific facial attributes were manually created before investigating how accurately the K-Means algorithm can cluster the resulting embeddings based on attribute discrimination.
The features examined in this work are categorised as extra class and intra class. Extra class features are facial attributes that are distinguishable between different identities, namely gender, skin tone and age. Intra class features are noted as facial attributes that distinguish between one unique identity, namely emotions, the presence and absence of beards, and the presence and absence of glasses. 
The images used for the test datasets were selected from the \textit{facial expressions} dataset available from the Muxspace GitHub repository \cite{Brian2016}. The \textit{facial expressions} dataset consists of 13,718 unprocessed images and was chosen due to its diversity in terms of gender, age, and ethnicity. It also displays variety in terms of emotions for unique identities, making it suitable for some intra class tests. 
To perform intra class attribute testing, each dataset must contain only one unique identity. However, for some intra class tests, the \textit{facial expressions} dataset did not contain enough images of the same identity with specific features. For example, as this dataset was created for facial expression diversity, finding images of the same identity with/without beards was challenging. Therefore, images were manually sourced from the Internet to create the required datasets for tests in which the attributes of beards and glasses were examined. Furthermore, as creating these datasets can be a time-intensive task, simple data augmentations are applied to images used in the intra class tests to increase the sample size to that of the extra class tests. These augmentations are discussed in greater detail here \ref{subsec:intraClass}.

Images used for the extra class tests were manually chosen from the \textit{facial expressions} dataset. Images for the test datasets are manually chosen based on the desired attribute for discrimination. For example, in a test where the attribute gender is investigated, the test dataset would contain 100 images, of which 50 are male, and 50 are female. Each image is manually labelled based on the perceived gender of the person within the image.
   
Each image within the dataset is then passed through the model to create an embedding. The model applies the following steps prior to creating the embedding representation. The frontal face detector detects the face within the image and places a bounding box around it. This verifies that a face is present and that the image does not contain more than one face. The landmarks for the detected face are then identified using the \textit{shape predictor 68 face landmarks} model \cite{King2009}; these landmarks are used to precisely localise the face. The images and their respective landmarks are then passed to the \textit{dlib face recognition ResNet model v1} which converts the images into their respective 128-dimensional embeddings.

Once the embeddings have been created, clustering is performed on the embeddings using Scikit-learns K-Means algorithm \cite{Pedregosa2011}; this experimental process can be seen in Figure \ref{fig:figure1}. The Chinese whispers clustering algorithm from the Dlib library was used initially. However, this algorithm did not perform well and was substituted for the K-Means algorithm. Several tests were carried out to compare the initialisation of the K-Means algorithm using random seeds against manual seeds. The random seeds provided more consistent results; therefore, the rest of the tests were carried out using random seed initialisation. This form of K-Means initialisation can create slightly different clusters on occasion. Each dataset was run through the algorithm five times to obtain a more reliable average classification accuracy. 

\begin{figure}[t]
\centering
\includegraphics[scale=1]{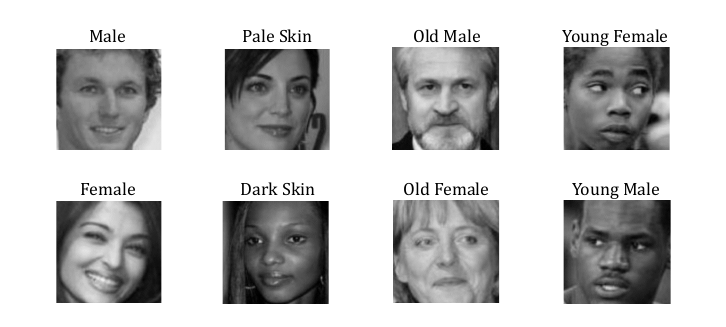}
\caption{Examples of images used for the initial stage of extra class testing, along with labels of how they were manually classified.}
\label{fig:figure2}
\end{figure}

When the data is clustered, output labels are generated by K-Means, which are then compared with the manually created labels to assess the clustering performance through confusion matrices.

{\section{Evaluation}\label{sec4}}
{\subsection{Extra Class}\label{subsec:extraClass}}
A set of experiments were undertaken to examine the extra class attributes representing gender, skin tone, and age. Examples of images used in these tests can be seen in Figure \ref{fig:figure2}. K-Means was initialised with random seeds and two clusters for each test. While it is appreciated that these attributes exist on a continuum, the attributes are classified in a binary fashion to understand how the representation within the embedding is structured. Therefore, dichotomic clusters are used as gender is classified as male/female, skin tone as dark/pale and age as young/old, where young is defined as individuals under the age of 50 and old as over 50.

The sample sizes for each test were 100 images, each test was run five times, and the average accuracy for each test, across all test runs, was 99.3\%, 99.3\% and 94.1\% for gender, skin tone and age, respectively. The average accuracy represents the percentage of images assigned to the correct cluster over the number of test runs (five for this experiment). It is evident from the high clustering accuracies achieved above that the extra class discriminative properties of gender, skin tone and age are represented within the embeddings. 

\begin{figure}[t]
\centering
\includegraphics[scale=0.6]{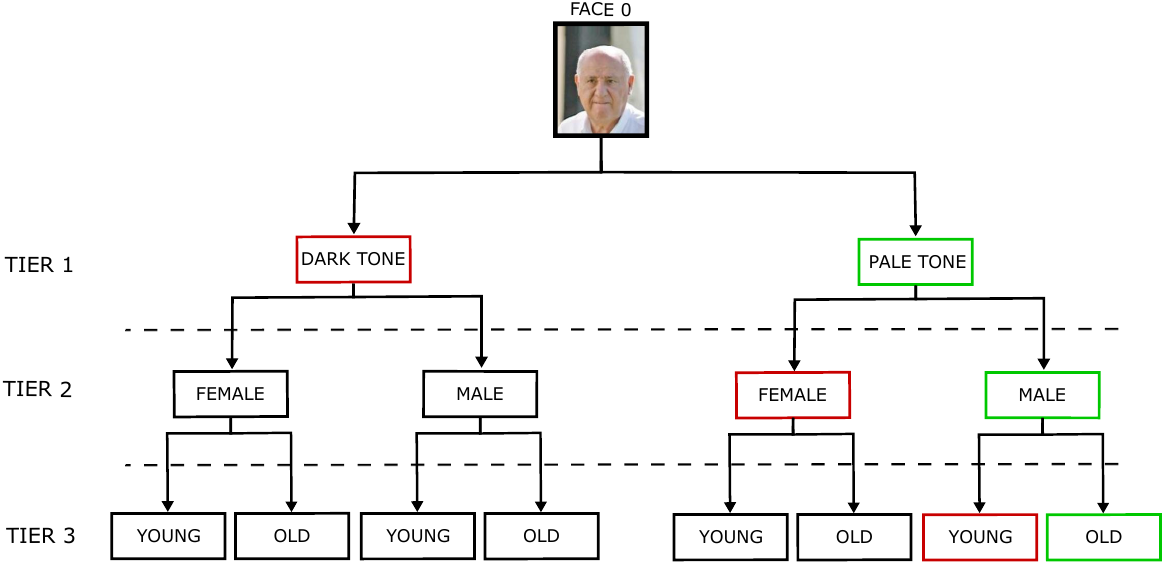}
\caption{Hierarchy of extra class attribute representation within the embedding structure. Tier 1 represents the attribute skin tone which takes priority within the structure, while Tier 3 represents the attribute age which ranks the lowest. The green boxes represent how face 0 was classified during the clustering process.}
\label{fig:figure3}
\end{figure}

An important outcome of this experiment was that it demonstrated the hierarchy of attributes within the embeddings. Each time the K-Means algorithm was run with two clusters, the embeddings would always cluster based on skin tone.
Therefore, to examine the attributes of gender and age, all images in the dataset needed to contain only one skin tone. This requirement demonstrates that a hierarchy of representation exists between features within the embedding structure, such that specific attributes have a stronger representation within the embedding structure.
A new dataset of 200 images was created and manually labelled to explore this condition further, ensuring an equal representation of attributes within the dataset. In the first part of this experiment, K-Means was initialised with two clusters to determine which attribute held prominence above the other two. This process was repeated for K-Means initialised with four and eight clusters. The results of these experiments can be seen in Table~\ref{tab:table1}. When using two clusters, the data was classified based on skin tone, such that one cluster contained pale-skinned people and the other contained dark-skinned people. When four clusters were used, the clusters formed around the attributes of skin tone and gender. Of the four clusters formed, one contained males with dark skin tone, a second cluster contained females with dark skin tone, a third contained females with pale skin tone and a fourth contained males with pale skin tone.
Finally, when eight clusters were used, each image was classified based on the skin tone, gender and age of the person in the image. For example, one cluster would contain old dark-skinned males, and another would contain young dark-skinned males. This extra class attribute representation hierarchy can be seen in Figure \ref{fig:figure3}. These results verify that not only are the extra class attributes of age, gender and skin tone represented within the embedding structure, but they also follow a hierarchy of representation whereby specific attributes are represented better or deemed more critical to the facial recognition task during training. These attributes rank in order of skin tone, gender, and age, respectively. Subsequently, this new information regarding the strong representation of facial attributes within the embedding structure allows for the sub-discrimination of these attributes without retraining the original DML model. It is recognised that the model does not perform as strongly for the attribute age compared to the attributes of gender and skin tone. We speculate that the attributes of gender and skin tone can be represented easier in a binary form and that this may not be the case for the attribute age. In future works, representing age in more distinct age groups, such as 0-5 years, 6-10 years etc., may prove beneficial rather than over 50 years and under 50 years.

\begin{table*}[ht]
\centering
\caption{Results of the $2^{nd}$ extra class test where the hierarchical representation of features in the embedding structure is examined. Outcomes indicate the ability to correctly identify a person's gender, age and skin tone with high classification accuracy (shown in bold). It is also evident that a hierarchical coherence exists between features in the embedding structure. The features representing the extra class attributes rank in order of skin tone, gender and age, respectively.}
\label{tab:table1}
\resizebox{\linewidth}{!}{%
\renewcommand{\arraystretch}{1.5}
\begin{tabular}{clcccccc}
\multicolumn{1}{l}{}                                                                      &                          & \multicolumn{1}{l}{} & \multicolumn{1}{l}{} & \multicolumn{1}{l}{} & \multicolumn{1}{l}{} & \multicolumn{1}{l}{} & \multicolumn{1}{l}{}                           \\ 
\hline
\multicolumn{1}{l}{\textbf{No. Clusters}}                                                 & \textbf{Cluster Content} & \textbf{Test 1}      & \textbf{Test 2}      & \textbf{Test 3}      & \textbf{Test 4}      & \textbf{Test 5}      & \multicolumn{1}{l}{\textbf{Average Accuracy}}  \\ 
\hline
\rowcolor[rgb]{0.961,0.961,0.961} {\cellcolor[rgb]{0.941,0.941,0.941}}                    & Dark Skin                    & 99.0\%               & 99.0\%               & 99.0\%               & 99.0\%               & 99.0\%               & \textbf{99.0\%}                                \\
\rowcolor[rgb]{0.961,0.961,0.961} \multirow{-2}{*}{{\cellcolor[rgb]{0.941,0.941,0.941}}2} & Pale Skin                    & 100.0\%              & 100.0\%              & 100.0\%              & 100.0\%              & 100.0\%              & \textbf{100.0\%}                               \\ 
\hline
{\cellcolor[rgb]{0.859,0.859,0.859}}                                                      & Dark Skin                    & 99.0\%               & 99.0\%               & 99.0\%               & 99.0\%               & 99.0\%               & 99.0\%                                         \\
{\cellcolor[rgb]{0.859,0.859,0.859}}                                                      & Pale Skin                    & 100.0\%              & 99.0\%               & 99.0\%               & 99.0\%               & 100.0\%              & 99.4\%                                         \\
\rowcolor[rgb]{0.961,0.961,0.961} {\cellcolor[rgb]{0.859,0.859,0.859}}                    & Male                     & 100.0\%              & 100.0\%              & 100.0\%              & 100.0\%              & 100.0\%              & \textbf{100.0\%}                               \\
\rowcolor[rgb]{0.961,0.961,0.961} \multirow{-4}{*}{{\cellcolor[rgb]{0.859,0.859,0.859}}4} & Female                   & 99.0\%               & 99.0\%               & 99.0\%               & 99.0\%               & 99.0\%               & \textbf{99.0\%}                                \\ 
\hline
{\cellcolor[rgb]{0.78,0.78,0.78}}                                                         & Male                     & 99.0\%               & 100.0\%              & 99.0\%               & 99.0\%               & 100.0\%              & 99.4\%                                         \\
{\cellcolor[rgb]{0.78,0.78,0.78}}                                                         & Female                   & 99.0\%               & 99.0\%               & 99.0\%               & 97.0\%               & 99.0\%               & 98.6\%                                         \\
{\cellcolor[rgb]{0.78,0.78,0.78}}                                                         & Dark Skin                    & 99.0\%               & 99.0\%               & 98.0\%               & 99.0\%               & 98.0\%               & 98.6\%                                         \\
{\cellcolor[rgb]{0.78,0.78,0.78}}                                                         & Pale Skin                    & 99.0\%               & 99.0\%               & 99.0\%               & 99.0\%               & 99.0\%               & 99.0\%                                         \\
\rowcolor[rgb]{0.961,0.961,0.961} {\cellcolor[rgb]{0.78,0.78,0.78}}                       & Young                    & 94.0\%               & 90.0\%               & 91.0\%               & 93.0\%               & 87.0\%               & \textbf{91.0\%}                                \\
\rowcolor[rgb]{0.961,0.961,0.961} \multirow{-6}{*}{{\cellcolor[rgb]{0.78,0.78,0.78}}8}    & Old                      & 88.0\%               & 79.0\%               & 89.0\%               & 80.0\%               & 80.0\%               & \textbf{83.2\%}                                \\
\hline
\end{tabular}
}
\end{table*}

\begin{figure}[t]
\centering
\includegraphics[width=1\textwidth]{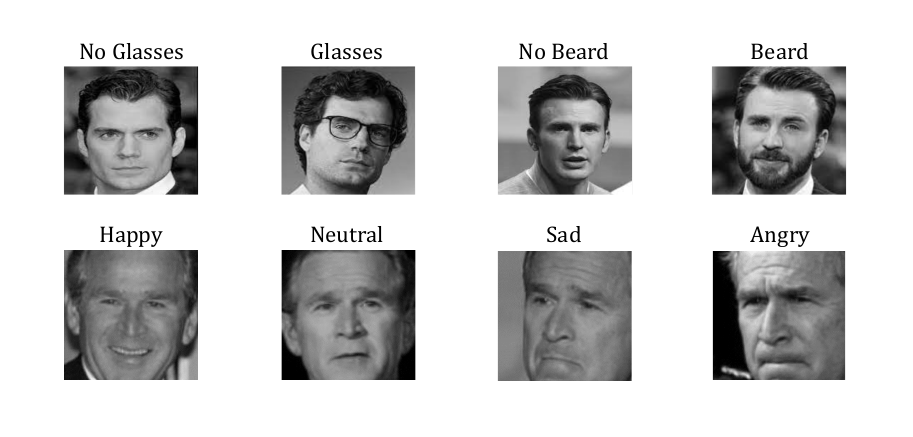}
\caption{Examples of images sourced from the Internet that were used to examine the presence of beards, glasses, and emotions as intra class discriminative properties.}
\label{fig:figure4}
\end{figure}

In the next section of this paper, we investigate the presence of the intra class attributes of beards, glasses, and emotions within the embedding structure, along with the possibility that the features representing the intra class attributes behave in the same manner as the extra class results indicate.

{\subsection{Intra Class}\label{subsec:intraClass}}
As discussed in Section~\ref{subsec:experiment procedure}, the \textit{facial expressions} dataset was compatible with the intra class experiments as it contained 3--10 images for each unique identity and showed significant variety in terms of emotions for each unique person. The initial intra class test mirrored the initial extra class test in that we first examined whether specific intra class attributes are represented within the embedding structure. Firstly, the four emotions happy, angry, sad, and neutral were investigated as these were the best-represented emotions in the \textit{facial expressions} dataset. However, although these emotions were represented the best, they were not represented sufficiently, resulting in a shortage of images and subsequently, very small datasets. To overcome this, basic image augmentations \cite{albumentations} were applied to bulk the sample size up to 80 images, near the 100 sample size for the extra class tests. The original datasets contained ten images, five images/attribute, and three augmentations were applied to the original images. Firstly, the original images are rotated at 5 degrees resulting in 20 total images. Secondly, Gaussian blur is applied to all images resulting in 40 total images. Finally, image compression is applied to all images, resulting in 80 images. Each attribute combination test was run ten times; the average accuracy and max accuracy across all 10 test runs can be seen in Table~\ref{tab:table2}. High maximum accuracies for most emotion combinations suggest that emotions are represented within the embedding structure. However, low average accuracies suggest that one-shot or few-shot learning techniques may be required to improve the sub-discriminative task. Additionally, the low average accuracies could result from applying an objective metric on a subjective attribute, such that the labels provided by the creators of the \textit{facial\_expressions} dataset can be subjective and may not be entirely representative of the emotion, resulting in lower classification accuracies. An example of this can be seen in \ref{fig:figure4}; it would be hard for a human to correctly label the pictures of angry and sad George W. Bush as both images are very similar in facial expression.

In the remainder of this section, we investigate whether the intra class attributes of beards and glasses are represented within the embedding structure.

\begin{table}[]
\caption{Results of the initial Intra class test, where the presence of the attribute emotions are examined. Average refers to the average accuracy across 10 test runs, while max refers to the maximum accuracy achieved across 10 test runs. Four separate identities (0 - 3) are used across all tests. The highest average accuracy is in bold.}
\label{tab:table2}
\begin{adjustbox}{width=1\textwidth}
\renewcommand{\arraystretch}{1.5}
\begin{tabular}{llcccc}
\hline
\textbf{Emotion 1} & \textbf{Emotion 2} & \textbf{Identity} & \textbf{Sample Size} & \textbf{Average} & \textbf{Max} \\ \hline
Anger              & Sad                & 0                 & 80                   & 60.0\%             & 90.0\%           \\
Anger              & Neutral            & 0                 & 80                   & \textbf{86.0\% }             & 100.0\%           \\
Anger              & Happy              & 0                 & 80                   & 65.0\%               & 90.0\%           \\
Happy              & Neutral            & 1                 & 80                   & 50.0\%               & 70.0\%           \\
Happy              & Sad                & 2                 & 80                   & 64.0\%               & 80.0\%           \\
Neutral            & Sad                & 3                 & 80                   & 46.0\%               & 60.0\%           \\ \hline
\end{tabular}
\end{adjustbox}
\end{table}

\begin{table}[]
\caption{Results of the $2^{nd}$ Intra class test, where the presence of the attributes beards and glasses are examined. Average refers to the average accuracy across 10 test runs, while max refers to the maximum accuracy achieved across 10 test runs. Five separate identities (0 - 4) are used across all tests. The highest average accuracy for each attribute is in bold.}
\label{tab:table3}
\begin{adjustbox}{width=1\textwidth}
\renewcommand{\arraystretch}{1.5}
\begin{tabular}{llcccc}
\hline
\textbf{Attribute Test} & \textbf{Identity} & \textbf{Sample Size} & \textbf{Average} & \textbf{Max} \\ \hline
\multirow{3}{*}{Beard vs No Beard}                & \multicolumn{1}{c}{0}                 & 80                   & 52.0\%               & 60.0\%           \\
           & \multicolumn{1}{c}{1}                  & 80                   & \textbf{90.0\% }             & 100.0\%           \\
            & \multicolumn{1}{c}{2}                  & 80                   & 85.0\%               & 100.0\%           \\ \hline
\multirow{3}{*}{Glasses vs No Glasses}            & \multicolumn{1}{c}{2}                  & 80                   & 70.0\%               & 100.0\%           \\
               & \multicolumn{1}{c}{3}                  & 80                   & 70.0\%               & 100.0\%           \\
              & \multicolumn{1}{c}{4}                  & 80                   & \textbf{76.0\% }             & 90.0\%           \\ \hline
\end{tabular}
\end{adjustbox}
\end{table}

\begin{figure}[t]
\centering
\includegraphics[width=1\textwidth]{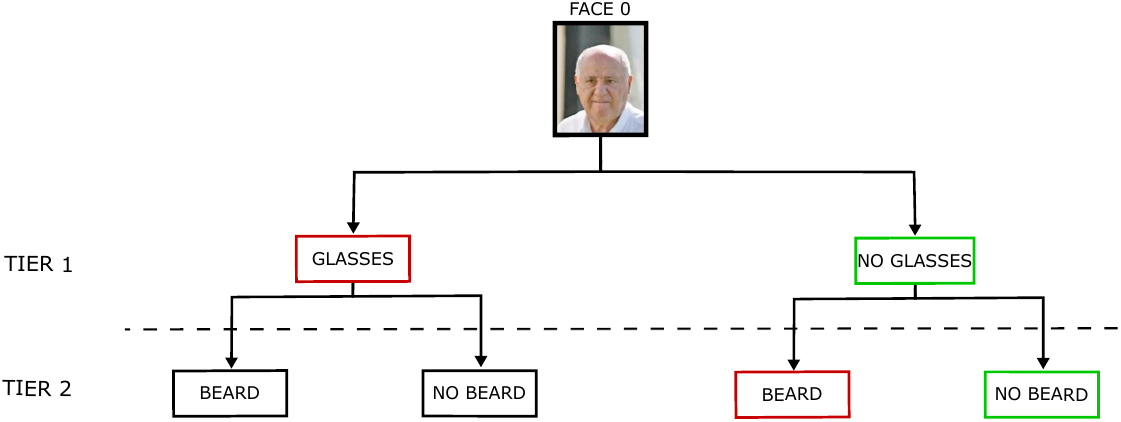}
\caption{Hierarchy of intra class attribute representation within the embedding structure. Tier 1 represents the attribute glasses which take priority within the structure, while Tier 2 represents the attribute beards which rank the lowest. The green boxes represent how face 0 was classified during the clustering process.}
\label{fig:figure5}
\end{figure}

The presence of beards and glasses was examined using three separate datasets per discriminative property. None of the datasets used in prior experiments contained enough images of the same identity with/without beards/glasses; therefore, images were manually sourced from the Internet; examples of these images can be seen in Figure~\ref{fig:figure4}.   
Each dataset contains ten original images, augmented in the same manner as the initial intra class test to increase the sample size to 80 images; results from these experiments are shown in Table \ref{tab:table3}. High maximum and average accuracies for most tests suggest that the attributes of beards and glasses are represented within the embedding structure. Similar to the extra class results, the intra class attributes of beards and glasses follow a hierarchy of representation within the embedding structure such that glasses rank above beards; this hierarchy can be seen in Figure \ref{fig:figure5}. The attributes of beards and glasses have a more substantial representation within the embedding structure in comparison to the attribute emotions; this is highlighted by higher average and max classification accuracies in most cases. The attribute glasses display more stable results as there is little variation between max and average accuracies for each dataset. Furthermore, although the attribute beards obtained the highest average accuracy of 90\%, the low average max and average accuracies for identity 0 suggest the attribute glasses may have a stronger or more stable representation within facial recognition embeddings.

{\subsection{Summary of Intra and Extra class tests}\label{subsec:summary_results}}
In summary, from initial inspection, the extra class attributes of skin tone, gender, and age are represented within the structure of facial recognition embeddings. These facial attributes are deemed essential features by the \gls{dml} model during the training of a facial recognition model. Additionally, results suggest that a hierarchical coherence exists between attributes within the embedding structure whereby specific attributes are better represented; for example, when using dichotomic clusters, if a dataset contained more than one skin tone, the embeddings would always cluster based on skin tone. This property tells us that certain features, in our case, the facial attribute of skin tone, best differentiate between different identities. Furthermore, the intra class attributes of beards and glasses have been identified to also exist within the embedding structure.
Additionally, results indicate that the intra class attributes of emotions are not represented as strongly as other attributes. The model under investigation performs facial recognition and aims to differentiate between different identities. Therefore, we speculate that the attribute emotion is not deemed as an essential feature by the model for the facial recognition task as this is not a typical attribute used to differentiate between identities in comparison to skin tone, gender, age and, in some cases, beards or glasses. Overall, most intra and extra class attributes examined in this study were found to hold representation within the embedding structure. Subsequently, the presence of these attributes allows for the sub-discrimination between features not presented during the initial training of the DML model.

\section{Application of Results}\label{sec5}
While the results of this study suggest that additional techniques are required for the application of intra class discriminative properties, a large avenue appears prevalent for the application of extra class discriminative properties. One possible avenue includes training one of the many existing unsupervised learning algorithms to cluster data based on skin tone, gender and age. This can be accomplished by saving the resulting cluster centroids in training and clustering unseen data by taking the Euclidean distance of each unseen embedding and classifying its discriminative properties based on which cluster centroid it lies closest to. To highlight the possibility of this application, a dataset consisting of 1000 samples, where each sample represents a unique identity, was manually created from the CelebA dataset \cite{DBLP:journals/corr/LiuLWT14}. The dataset used for this experiment consisted of the following:

\begin{itemize}
   \item 125 young pale-skinned males;
   \item 125 young dark-skinned males;
   \item 125 young pale-skinned females;
   \item 125 young dark-skinned females;
   \item 125 old pale-skinned males;
   \item 125 old pale-skinned females;
   \item 125 old dark-skinned males;
   \item 125 old dark-skinned females.
\end{itemize}

\begin{figure}[t]
\centering
\includegraphics[width=1\textwidth]{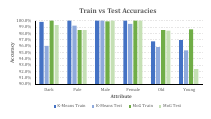}
\caption{The train and test accuracies for both \gls{mog} and K-Means algorithms. Both algorithms display high accuracies for each extra class attribute.}
\label{fig:figure6}
\end{figure}

This dataset was broken down into 70\% train data and 30\% test data; a validation set was not used as we are training an unsupervised learning algorithm. Following previous extra class experiments, K-Means initialised with random seeds was chosen as the unsupervised learning algorithm. The training data was processed through K-Means several times using two clusters initially, then four clusters and finally eight clusters. Each cluster's centroids are generated by the K-Means algorithm and saved. The centroids that produced the highest attribute accuracies in training are chosen to cluster the test data. Classification is achieved by classifying embeddings based on the cluster centroid they lie closest to. The results of this experiment can be seen in Figure~\ref{fig:figure6} denoted as 'K-Means Train' and 'K-Means Test'. 

The results from this experiment indicate the possibility of using means generated by an unsupervised learning algorithm to cluster embeddings based on unseen extra class discriminative properties. This yields an advantage in terms of computation speed for the sub-discrimination task. In contrast to comparing a new embedding to each known identity, it is more computationally efficient to compare a new embedding to each cluster centroid and then compare the new embedding to the known identity in that cluster. This comparison reduction, especially with larger sample sizes, can significantly increase computation speed while maintaining high performance on the sub-discrimination task, as shown by the high attribute accuracies achieved for this experiment.

In the final stage of this experimental study, the substantial drop in accuracy for the dark skin attribute between the train and test sets in the prior experiment is examined. To investigate whether K-Means was accurate enough, the \gls{mog} unsupervised algorithm was trained and tested on the same datasets. The method used to initialise the weights, the means and the precision was left as K-Means, the default initial parameter option for the \gls{mog} algorithm. 

Figure~\ref{fig:figure6} shows the train and test accuracies for both \gls{mog} and K-Means. The results show that both algorithms can accurately identify extra class discriminative properties. \gls{mog} does increase the accuracy of the dark skin attribute by 3.3\%. However, it does not perform as well as K-Means for the pale skin attribute dropping by 0.7\%. Although both algorithms display good performance, the results are not conclusive enough to state which algorithm performs best. A path for future work in the area is to examine the behaviour of several unsupervised algorithms to determine which performs best in terms of feature classification.

\begin{figure}[t]
\centering
\includegraphics[width=1\textwidth]{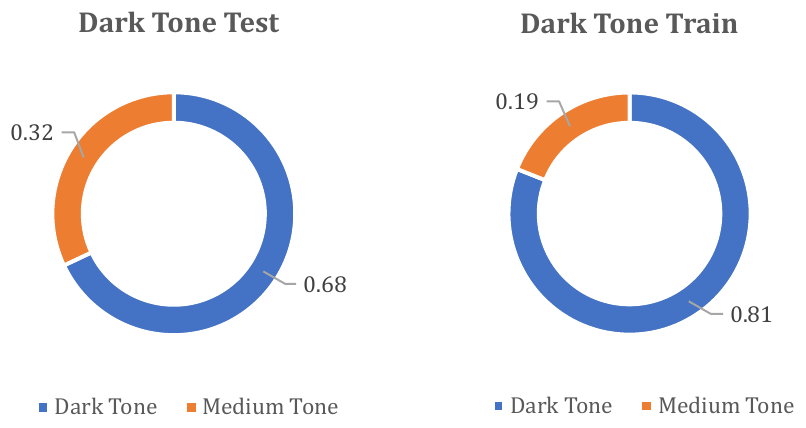}
\caption{The contents of the train and test sets for the dark skin tone attribute. An uneven representation of dark and medium tone images between the test and train sets could explain the decrease in accuracy for the dark skin attribute.}
\label{fig:figure7}
\end{figure}

The decrease in accuracy for the dark skin attribute can be seen as a consequence of the content of the test and training datasets. Additionally, due to the limitation of the experiment whereby, this study attempts to perform binary classification on continuous attributes. For the chosen dataset, dark skin people are defined as having dark or medium skin tones. It is noted that 19\% of the images representing the dark class in the training set were medium skin tone, while 32\% of the images representing the dark class in the test set were medium skin tone. The content of the train and test sets for the dark skin tone attribute can be seen in Figure \ref{fig:figure7}. This indicates that because of the low percentage of medium skin tone images in the training set, the model is less likely to classify medium skin tone data in the test dataset correctly; therefore, providing a more balanced training set in terms of skin tone should lead to improved results.

\section{Future Work}\label{sec6}
The most relevant area concerning future work is applying the techniques developed in this study to embeddings created by other data.
Embeddings created for any purpose may contain inherent information that could be used to perform sub-discriminative tasks through near zero-shot learning. Speech recognition is one of many possible areas where these principles can be applied. For example, spectrograms of voices can be represented as embeddings, and through the use of an unsupervised learning algorithm, it may be possible to identify disparate speakers accurately. Utilising more robust and clearly labelled datasets in future experiments could improve results and produce more sound conclusions. Setting rigid boundaries on continuous attributes may have negatively affected  impacted results, although the theory behind this work holds true, despite the limitations of performing binary classification on continuous attributes. A more in-depth computational analysis, alongside the deployment of the lightweight framework discussed in \ref{sec5} on network edge devices, could aid in the development of low-cost, low-power-consuming facial recognition systems.

\section{Conclusion}\label{sec7}
This paper experimentally evaluated the presence and application of additional information represented within the structure of embeddings generated for facial recognition. Results confirm the presence of intra and extra class facial attributes within the embedding structure. In addition, it is shown that through the use of clustering, this inherent information has application in sub-discriminating additional features not presented during the initial training of the \gls{dml} model. Extra class sub-discrimination can be achieved with high accuracy, notably with an average attribute accuracy of 99.3\%, 99.3\% and 94.1\% for skin tone, gender, and age, respectively.
In addition, the intra class attributes of beards and glasses produce an average attribute accuracy of 90.0\% and 76.0\%, respectively. It is important to note that this study has some limitations that should be acknowledged. Firstly, the sample sizes used in this study are quite small, which may limit the generalizability of the results. Secondly, this study only tested one specific model for facial recognition, and the results may not necessarily apply to other models outside of facial recognition. Finally, while the results show promising potential for the presence and use of additional feature information in \gls{dml}, further research is needed to validate and expand upon these findings. The main findings of this experimental study are summarised below:
\begin{itemize}
    \item It is possible to perform extra class sub-discriminative tasks with a high degree of accuracy through the use of unsupervised clustering algorithms. The discovery of inherent information within embeddings designed for facial recognition confirms the ability to perform extra class sub-discrimination, namely for the attributes gender, skin tone and age;
    \item The results from  the intra class sub-discrimination experiments highlight that the attribute of emotions may not be considered an important feature by the \gls{dml} model, although the attributes of beards and glasses seem to have some importance in the facial recognition task;
    \item A hierarchical coherence between attributes suggests that specific facial attributes take precedence during the training/creation of facial recognition embeddings; our results indicate that the extra class attributes rank in order of skin tone, gender, and age, respectively, and the intra class attributes rank in order of glasses and beards, respectively.
    \item Finally, the possibility of training unsupervised algorithms to perform extra class sub-discrimination at extremely high accuracies by saving cluster centroids created during training is demonstrated. Additionally, this method can reduce embedding comparisons for the facial recognition task, reducing computation times and the power consumption of facial recognition systems. 
\end{itemize}

\bibliography{mybibfile.bib}

\end{document}